\PassOptionsToPackage{dvipsnames, table, xcdraw}{xcolor}
\documentclass[twocolumn]{mc_nankai}

\usepackage{utfsym}
\usepackage{booktabs}
\usepackage{amsmath, amssymb, mathtools, bm}
\usepackage{multirow, enumitem, subcaption, float, wrapfig}
\usepackage{pifont}
\usepackage{overpic}

\usepackage[most]{tcolorbox}
\usepackage{lipsum} 
\usepackage{titlesec}
\usepackage{bm}
\usepackage{makecell}

\titlespacing*{\subsection}
  {0pt} 
  {1.5ex plus 0.5ex minus .2ex} 
  {0.8ex plus .2ex} 

\titlespacing*{\section}
  {0pt} 
  {1.5ex plus 0.5ex minus .2ex} 
  {0.8ex plus .2ex} 

\renewcommand{\myPara}[1]{\vspace{5pt}\noindent\textbf{#1}}
\definecolor{mygray}{HTML}{E7E6E6}

\newcommand{\cmark}{\ding{51}} 
\newcommand{\xmark}{\ding{55}} 

\newtcolorbox{mybox}[1][\linewidth]{
    colback=gray!15,
    colframe=black!60,
    arc=3mm,
    boxrule=1pt,
    drop shadow={black!50!white},
    width=#1,  
}

\definecolor{myredbg}{RGB}{253, 232, 232}
\definecolor{mypurplebg}{RGB}{236, 232, 246}
\definecolor{myredhl}{RGB}{220, 50, 50}
\definecolor{myyellowhl}{RGB}{255, 255, 0}

\newcommand{\redhighlight}[1]{%
  \tcbox[on line, boxsep=1pt, left=0pt, right=0pt, top=0pt, bottom=0pt, colback=myredhl, coltext=white, boxrule=0pt]{#1}%
}
\newcommand{\yellowhighlight}[1]{%
  \tcbox[on line, boxsep=1pt, left=0pt, right=0pt, top=0pt, bottom=0pt, colback=myyellowhl, boxrule=0pt]{#1}%
}

\newtcolorbox{fullwidthsidebysidebox}[2][]{%
    enhanced,
    breakable,
    width=\textwidth, 
    colframe=#2!75!black,
    colback=#2,
    arc=3mm,
    boxrule=1pt,
    sidebyside,
    sidebyside gap=12pt,   
    lower separated=false,
    righthand width=0.65\textwidth,
    #1
}

\newcommand{\textsupsub}[3]{%
    #1$^{\text{#2}}_{\text{#3}}$%
}
\newcommand{\textsup}[2]{%
    #1$^{\text{#2}}$%
}
\newcommand{\textsub}[2]{%
    #1$_{\text{#2}}$%
}
\def\Ours{GlimpsePrune}

\definecolor{iccvblue}{rgb}{0.21,0.49,0.74}

\title{A Glimpse to Compress: Dynamic Visual Token Pruning for Large Vision-Language Models}

\author[1,2]{Quan-Sheng Zeng}
\author[1]{Yunheng Li}
\author[3]{Qilong Wang}
\author[4]{Peng-Tao Jiang}
\author[2]{Zuxuan Wu}
\author[1]{Ming-Ming Cheng}
\author[1]{Qibin Hou$^{\dagger}$}

\affiliation[1]{VCIP, CS, Nankai University}
\affiliation[2]{Shanghai Innovation Institute}
\affiliation[3]{Tianjin University}
\affiliation[4]{vivo Mobile Communication Co., Ltd}
\affiliationenter[]{$^{\dagger}$Corresponding author.}

\abstract{
Visual token compression is critical for Large Vision-Language Models (LVLMs) to efficiently process high-resolution inputs.
Existing methods that typically adopt fixed compression ratios cannot adapt to scenes of varying complexity,
often causing imprecise pruning that discards informative visual tokens and results in degraded model performance.
To address this issue, we introduce a dynamic pruning framework, GlimpsePrune, inspired by human cognition.
It takes a data-driven ``glimpse'' and prunes irrelevant visual tokens in a single forward pass before answer generation.
This approach prunes 92.6\% of visual tokens while on average fully retaining the baseline performance on free-form VQA tasks.
The reduced computational cost also enables more effective fine-tuning:
an enhanced GlimpsePrune+ achieves 110\% of the baseline performance while maintaining a similarly high pruning rate.
Our work paves a new way for building more powerful and efficient LVLMs.
}

\mclink[Project Page]{\url{https://github.com/HVision-NKU/GlimpsePrune}}

\begin{document}
\maketitle
\justifying

\section{Introduction}


The rise of Large Language Models (LLMs) has spurred the development of large vision language models (LVLMs).
Standard LVLMs encode visual inputs into tokens, which are then projected into the textual embedding space and concatenated with text instructions for an LLM to process.
A recent trend towards a fixed patch size, rather than a fixed number of patches, means that higher-resolution inputs often generate a large number of visual tokens.

\begin{figure}[t]
    \centering
    \includegraphics[width=\linewidth]{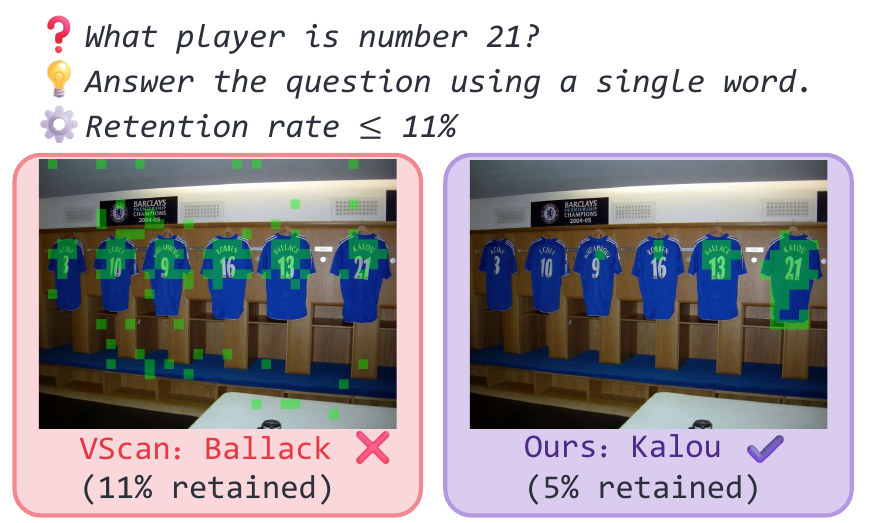}
    \caption{
        Compared to previous methods with fixed retention rates~\cite{zhang2025vscan}, 
        our method can dynamically select effective visual tokens while being more accurate.
    }
    \label{fig:redundancy}
\end{figure}

This proliferation of visual tokens creates a significant computational burden, as the number of visual tokens often dwarfs the textual query tokens.
Usually, only a small, query-dependent subset of these visual tokens is relevant for answering a given question.
This highlights the need for dynamic, instruction-guided visual token pruning. 
Existing methods either fix the compression ratio in the architectural design~\cite{chen2024llavolta,li2024llamavid} 
or use hand-crafted compression metric design~\cite{chen2024fastv,yang2025visionzip,zhang2025vscan}.
As shown in \figref{fig:redundancy}, when setting the same upper limit on the retention rate of visual tokens,
existing methods still retain a high degree of redundancy among the selected tokens. 
This motivates our central research question: How can we learn a data-driven metric for dynamic and efficient visual token pruning?

\begin{figure}[t]
    \centering
    \includegraphics[width=\linewidth]{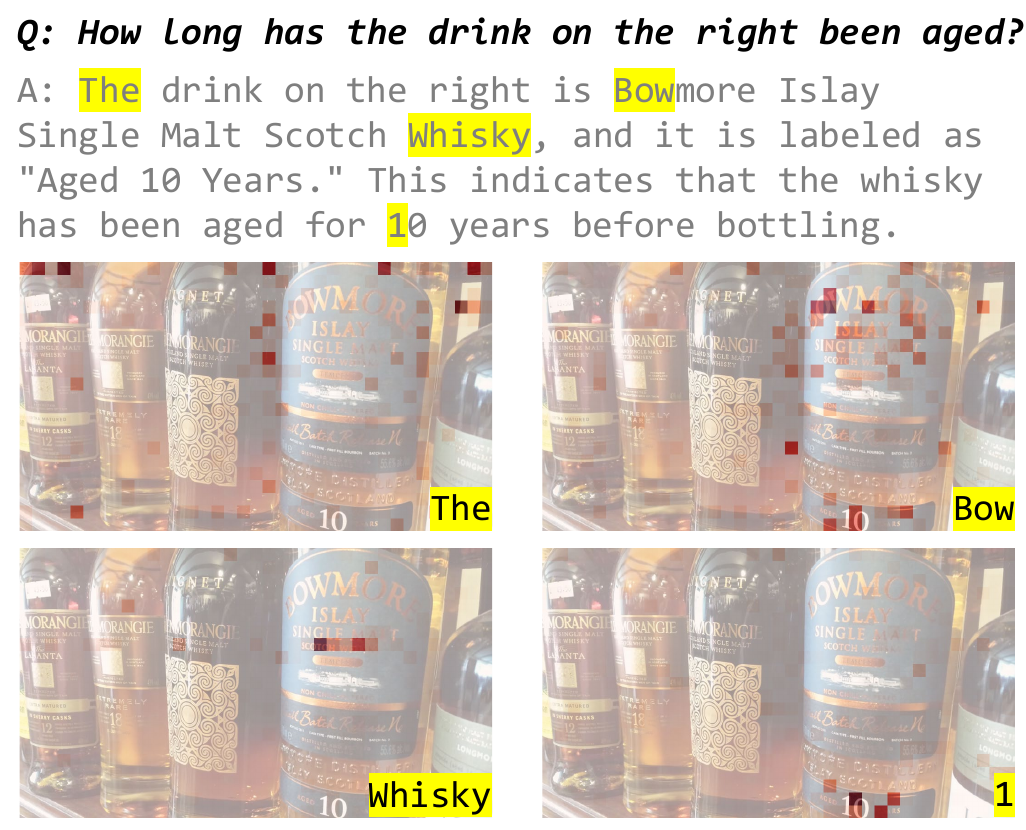}
    \caption{
        Without a brief prompt, the cross-attention map may initially fail to focus on relevant visual tokens, 
        and only begins to shift its focus once the generated text aligns with those visual elements.
    }
    \label{fig:seq_attn}
\end{figure}

Beyond their inflexible designs,
some existing token pruning methods exhibit a critical limitation in real-world, free-form response scenarios. 
This is because they operate on the flawed assumption that 
the cross-attention from the first generated token to the visual tokens reliably indicates visual token importance~\cite{chen2024fastv,zhang2025mllmsknowwhere}.
We find this assumption to be fragile.
\tabref{tab:brief_or_not} provides quantitative evidence, showing that the performance of prior methods drops sharply when they are not prompted for concise answers.
The reason is visually evident in \figref{fig:seq_attn}: 
when generating free-form responses, the cross-attention initially focuses on irrelevant regions, 
leading to imprecise pruning and a degraded final output.

\begin{table}[ht]
    \centering
    \small
    \setlength{\tabcolsep}{4pt}
    \begin{tabular}{ccccc}
    \toprule
    \thead{Method} & \thead{Qwen2.5-VL-7B} & \thead{PDrop} & \thead{VScan} & \thead{\Ours} \\
    \midrule
    w/ brief  &     0.936    & 0.753 & 0.845 & 0.929   \\
    w/o brief &     0.927    & 0.406 & 0.781 & 0.939   \\
    \bottomrule
    \end{tabular}
    \caption{
        On TextVQA, previous methods (PDrop, VScan) show high sensitivity to the brief prompt, 
        while the base model (Qwen2.5-VL-7B) and our method remain robust.
    }
    \label{tab:brief_or_not}
\end{table}

While dynamic pruning at each decoding step appears to be a straightforward solution, 
its practical utility is limited. 
The critical information for pruning
— the cross-attention maps that best indicate visual token importance —
has been shown to emerge only at deeper layers $K$ of the LLM decoder~\cite{chen2024fastv,xing2024pyramiddrop,zhang2025vscan}.
A per-step pruning strategy would thus have to operate after the forward pass through these initial $K$ layers,
which necessitates retaining the full KV cache for all visual tokens up to that point.
Consequently, this strategy would offer limited savings on computation and memory.

To address these challenges, we introduce \Ours{}, a dynamic visual token pruning framework 
inspired by the human cognitive process of first glancing at relevant areas before responding.
During the prefilling stage, we temporarily insert a learnable ``glimpse token'' after the instruction tokens.
We leverage the forward pass through the initial $K$ decoder layers to extract the cross-attention scores
between this glimpse token and all visual tokens,
and feed them into a lightweight predictor.
Both the glimpse token and the predictor are trained on a small amount of Grounded Question Answering (GQA) data
to learn a data-driven pruning metric.
Based on this, we perform a one-time pruning of irrelevant visual tokens 
and their corresponding KV cache entries across all model layers. 
This strategy reduces the computational and memory footprint for the remaining $L-K$ prefill layers 
and the entire subsequent decoding phase.


Extensive experiments demonstrate our predictor's strong generalization capabilities.
Integrated with Qwen2.5-VL-7B, \Ours{} achieves an average visual token pruning rate of 92.6\% while retaining 100\% of the original model's performance on free-form VQA tasks. 
Leveraging this efficiency, we show that under a reinforcement learning framework, our fine-tuned version, \Ours{}+, reaches 110\% 
of the original performance while maintaining a high pruning rate, highlighting a powerful synergy between our pruning strategy and advanced fine-tuning.

\begin{figure*}[t]
    \centering
    \includegraphics[width=0.95\linewidth]{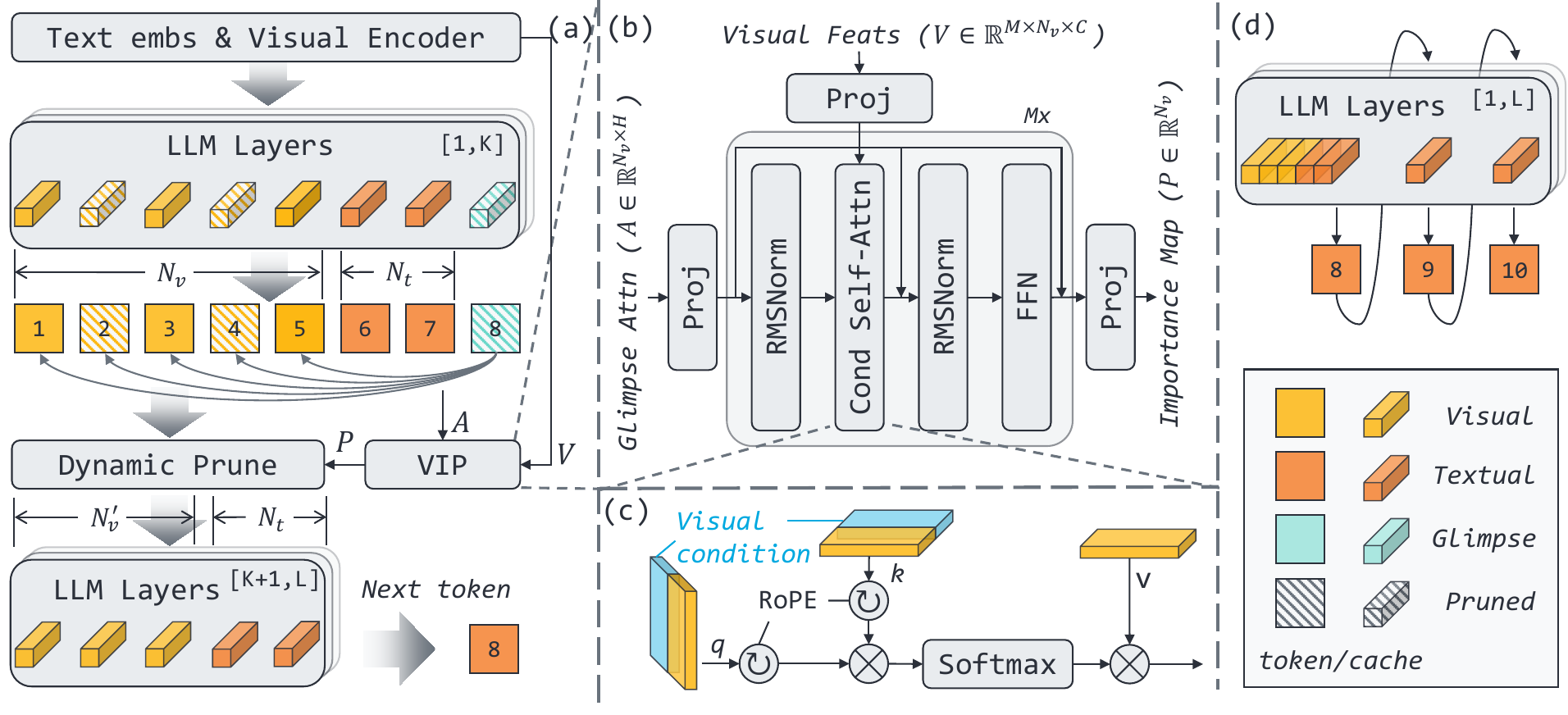}
    \caption{
        Overview of \Ours.
        (a) At the prefilling stage, we prune irrelevant visual tokens after $K$ layers of the LLM decoder.
        (b) The architecture of visual token importance predictor (VIP).
        (c) The details of conditional self-attention with 2D RoPE.
        (d) At the decoding stage, the answer generation relies on the pruned KV cache, saving memory and I/O bandwidth.
    }
    \label{fig:framework}
\end{figure*}

\section{Related Work}

\myPara{Large vision-language models.}
Pioneered by models like BLIP-2~\cite{li2023blip} and LLaVA~\cite{liu2023llava}, 
LVLMs combine a vision encoder, a projector, and an LLM to achieve visual understanding. 
Recent models, including the InternVL~\cite{chen2024internvl,chen2024far,wang2024mpo,chen2024expanding,zhu2025internvl3} 
and Qwen-VL~\cite{Qwen-VL,Qwen2-VL,Qwen2.5-VL} series, support dynamic, high-resolution inputs.
While enabling fine-grained analysis, this capability dramatically increases visual tokens, 
creating a computational bottleneck that motivates token compression research.

\myPara{Visual token compression.}
To mitigate the high computational cost of LVLMs, various token reduction strategies have been proposed. 
Some methods embed compression directly into the model architecture~\cite{chen2024llavolta,li2024llamavid}. 
However, these approaches are inflexible, suffering from fixed compression ratios tied to the architecture and requiring costly, full-scale retraining.
Consequently, training-free, post-hoc methods that prune or merge tokens have gained prominence,
which typically rely on two types of metrics: importance-based metrics and similarity-based metrics.

For importance-based metrics, prior works typically compute importance scores using the cross-attention 
from a global query representation to all visual tokens.
This query representation can be derived from the instruction token~\cite{han2024free,wang2024retake,xing2024pyramiddrop,zhang2025cdpruner,zhang2025vscan}, 
or a dedicated global visual token~\cite{yang2025visionzip,zhang2025vispruner,dhouib2025pact}.
However, such a hand-crafted importance metric can be unreliable for free-form generation tasks, 
where the user's core objective is not explicitly stated in the query, 
or the target information does not appear at the beginning of the generated response.
Similarity-based metrics are also used by some prior works to prune and merge visual tokens,
which typically use cosine distance or self-attention scores within the visual features~\cite{yang2025visionzip,alvar2025divprune,zhang2025cdpruner,yang2025topv,sun2025llava}.
However, these techniques incur substantial memory overhead from the computation of dense similarity matrices,
potentially negating the efficiency gains from optimizations like FlashAttention~\cite{dao2022flashattention,dao2023flashattention2}.

Our work presents a data-driven pruning metric to enable a dynamic compression ratio and addresses the limitations of hand-crafted ones.
Although some recent works also share the motivation for dynamic token compression~\cite{jin2024chat,huang2025dynamicllava,tao2025dycoke,liang2025dyrate},
our contribution is distinct in several key aspects.
(i) We identify and address the inadequacy of prior methods for free-form VQA tasks;
(ii) Our proposed framework employs a glimpse token and a lightweight predictor, which are supported by a tailored training methodology.
(iii) Our pruning strategy is fundamentally more efficient, performing a single full-depth KV cache prune per response. 
This is in stark contrast to the incremental pruning~\cite{chen2024fastv} or partial~\cite{xing2024pyramiddrop,zhang2025vscan} pruning in prior art, thus maximizing memory savings during decoding.

\section{Preliminaries}
Inference in LVLMs is a two-phase process: prefilling and decoding, facilitated by a KV cache. 
To mitigate the high memory cost of attention, techniques like FlashAttention~\cite{dao2022flashattention,dao2023flashattention2} are now standard.
We denote the visual and text token sequences by lengths $N_v$ and $N_t$, respectively (where typically $N_v \gg N_t$), 
and consider an LLM with $L$ decoder layers of hidden dimension $D$.
The prefilling phase is compute-intensive. It processes the initial $N_v + N_t$ tokens via causal self-attention, 
with a computational cost of $\mathcal{O}((N_v + N_t)^2D)$, to populate a KV cache of size $2L(N_v + N_t)D$. 
Conversely, the decoding phase is memory-bandwidth-intensive. It generates tokens autoregressively, 
where the attention cost is reduced to $\mathcal{O}((N_v + N_t)D)$ per token by leveraging the cache. 
The bottleneck thus shifts to the I/O cost of loading this large cache from memory.

While optimizations like FlashAttention reduce prefilling memory by avoiding the materialization of the full $\mathcal{O}((N_v + N_t)^2)$ attention matrix, 
this efficiency gain is lost by pruning methods that require this exact matrix for token similarity assessment~\cite{zhang2025cdpruner,zhang2025vscan}. 
In contrast, computing cross-attention from a single query to all visual tokens is a linear-cost operation compatible with FlashAttention's memory efficiency.

Our approach, therefore, intervenes mid-prefilling. 
After a designated layer $K$, we employ a lightweight selection module, 
informed by a linear-cost glimpse attention and pre-computed visual features,
to prune the visual token set from $N_v$ down to $N'_v$. 
Critically, this action simultaneously prunes the KV cache for all preceding layers ($1$ to $K$). 
This single intervention reduces the computational load for the remaining $L-K$ prefill layers and, 
more importantly, slashes both memory and I/O costs for the entire subsequent decoding phase.
More details are presented in \secref{sec:method}.

\begin{figure*}[t]
    \centering
    \includegraphics[width=0.95\linewidth]{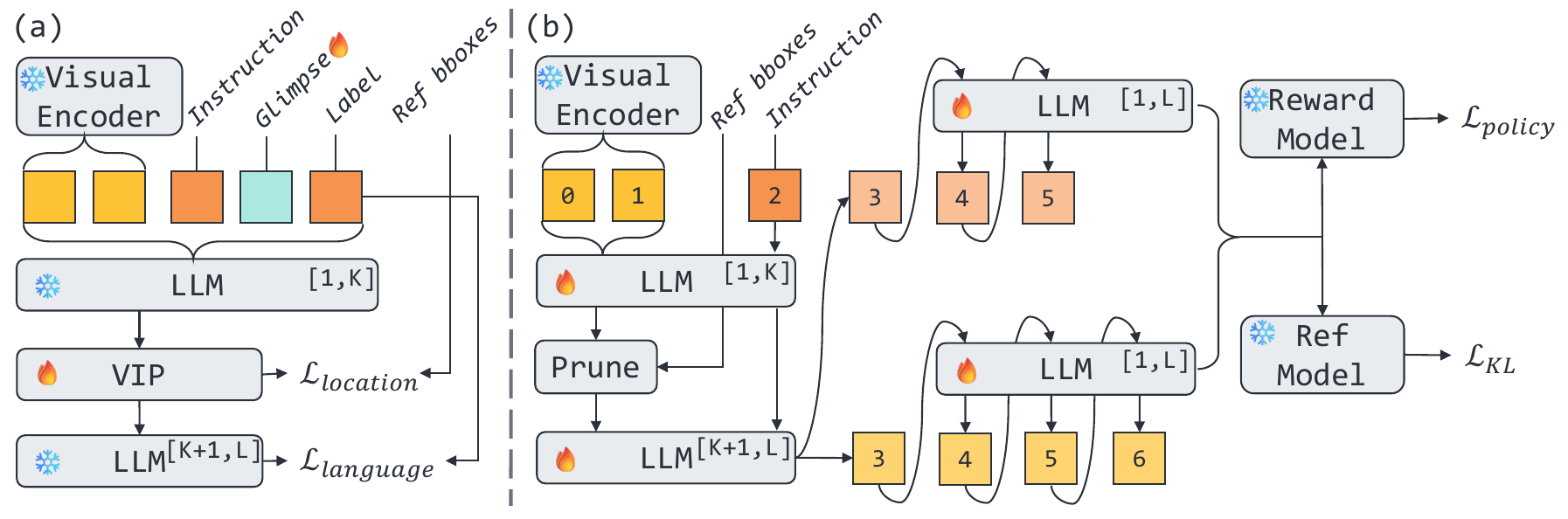}
    \caption{
    Overview of the training process:
        (a) Training glimpse token and VIP for data-driven pruning metric.
        (b) RL fine-tuning LLM for \Ours{}+.
    }
    \label{fig:training}
\end{figure*}

\section{Method}\label{sec:method}

\myPara{Hierarchical visual features and glimpse attention extraction.}
When the visual encoder processes a patched image into $N_v$ visual tokens,
we extract $C$-dimensional visual features, denoted as $V \in \mathbb{R}^{M \times N_v \times C}$,
from $M$ of its intermediate layers.
These hierarchical features provide multi-scale visual priors
that are crucial for the subsequent visual token importance prediction.


For the prefilling stage, we introduce an $L\times  D$ matrix of learnable embeddings as glimpse token.
The first embedding is concatenated to the end of the input sequence,
and then at each decoder layer, its respective embedding is added to the hidden state at that newly appended position.
The purpose of appending them at the final position is to leverage the LLM decoder's causal attention mechanism.
This allows the glimpse token to fully interact with both visual and textual tokens across all layers, 
thereby capturing early signals about the importance of visual tokens relevant to the user's instruction, 
all without perturbing the original processing of the visual and textual tokens themselves.

We posit that by an intermediate layer $K$, 
the glimpse token has already aggregated sufficiently clear importance information. 
Our subsequent experiments demonstrate that setting $K=\lceil2/3 \times L\rceil$ strikes a favorable trade-off 
between performance and the computational overhead of the prefilling stage. 
We then extract the \textit{glimpse attention} from the glimpse token to all visual tokens at layer $K$, 
denoted as $A \in \mathbb{R}^{N_v \times H}$, 
where $H$ is the number of attention heads.

\myPara{Visual importance prediction and one-shot pruning.}
Subsequently, the extracted glimpse attention $A$ and the multi-level visual features $V$
are fed into the selection module, named Visual Importance Predictor (VIP).
As illustrated in \figref{fig:framework}(b),
the VIP is composed of $M$ Self-Attention Blocks, 
equipped with 2D Rotary Position Embeddings (RoPE)~\cite{su2024roformer}.
The glimpse attention $A$ is first projected into a new embedding space $A' \in \mathbb{R}^{N_v \times E}$, 
where $E$ is the VIP's hidden dimension.
As illustrated in \figref{fig:framework}(c),
within the $m$-th self-attention block,
we project the corresponding $m$-th visual feature map $V_m$ to $V_m' \in \mathbb{R}^{N_v \times F}$
and concatenate it to the query and key vectors.
This allows the relationship within the visual tokens of $V$ to act as additional conditioning, 
influencing the final importance prediction. 

By setting the dimensions $E \ll D$, $F \ll C$ (where $D$ is the LLM's hidden dimension and $C$ is the visual feature dimension),
and implementing a memory-efficient attention mechanism, 
the computational overhead of the entire VIP remains negligible compared to a single LLM decoder layer.

The VIP's output, an importance map $P$, 
dynamically and precisely identifies visual tokens relevant to the user's instruction, 
adapting to different inputs and visual targets. 
Furthermore, a maximum retention ratio can be configured to sparsify the visual tokens 
even when $P$ focuses on large-scale visual objects.

Based on the importance map, we perform a \textit{one-shot pruning} at layer $K$. 
Unimportant visual tokens are removed from the layer-$K$ hidden state 
and their corresponding KV cache entries in all preceding layers,
permanently reducing the sequence length to $N'_v + N_t$ for all subsequent computations.
The glimpse token is also discarded so that the answer generation will not be influenced by it.

\myPara{Importance metric learning by glimpse token and VIP.}
The glimpse tokens and the VIP are trained while keeping the parameters of the base VLM frozen, 
using only a small amount of Grounded Question Answering (GQA) data. 
Each data sample consists of a question, its corresponding answer, 
and the bounding boxes of the visual regions relevant to the answer. 
Our training objectives are two-fold. 

On one hand, we apply a language modeling loss that encourages the model 
to generate the correct answer immediately after the position of the glimpse token. 
On the other hand, we employ a localization loss to optimize the VIP's output. 
This loss is a weighted sum of DiceLoss and Binary Cross-Entropy (BCE) Loss. 
  Given that bounding box annotations are often coarse, 
  we prioritize recalling foreground visual tokens over achieving precise boundary segmentation. 
  Therefore, we set the weight ratio for DiceLoss to BCE Loss at 10:1. 
  The language loss and the localization loss are balanced with a 1:1 weight ratio.

  Since the LVLM's original attention mechanism already possesses an inherent localization capability, 
  our training primarily serves to calibrate the newly introduced parameters and amplify this ability. 
  Consequently, our method exhibits strong generalization, a claim we will substantiate in the experimental section.

  \myPara{Reinforcement learning fine-tuning.}
  RL fine-tuning on dense visual inputs is computationally expensive and memory-intensive. 
  Combining with our pruning mechanism mitigates this cost by first employing a pruning mechanism to create a sparse visual context. 
  We then fine-tune the LLM decoder using the Group-wise Ranking Policy Optimization (GRPO) framework~\cite{shao2024deepseekmath}.

  As illustrated in \figref{fig:training}(b), the fine-tuning process begins after pruning at layer $K$. 
  The policy model generates multiple candidate responses, which are evaluated by a reward model to compute a policy loss. 
  The loss is regularized with a KL divergence term against a reference model.
  The new model, named \Ours{}+, is trained iteratively. We first update the LLM decoder via GRPO, and then train our glimpse token and predictor on the updated decoder. 
  This process allows \Ours{}+ to achieve superior performance over the baseline model even at a high pruning rate.

\begin{table*}[ht]
\centering
\setlength{\tabcolsep}{1pt}
\small
\begin{tabular}{lccccccccccccc}
\toprule
 &
  \multicolumn{2}{c}{\thead{General \\ VQA}} &
  \multicolumn{3}{c}{\thead{Relation \\ Reasoning}} &
  \thead{Fine \\ grained} &
  \multicolumn{5}{c}{\thead{Doc/Text}} &
  \thead{Chart} &
   \\
\cmidrule(lr){2-3} \cmidrule(lr){4-6} \cmidrule(lr){7-7} \cmidrule(lr){8-12} \cmidrule(lr){13-13}
\multirow{-2}{*}{\thead{Method}} &
  Flickr30k &
  V7W &
  GQA &
  OpenImages &
  VSR &
  CUB &
  DocVQA &
  TextCaps &
  TextVQA &
  DUDE &
  SROIE &
  InfoVQA &
  \multirow{-2}{*}{\thead{Average}} \\
\midrule
Qwen2.5-VL-7B &
  0.761 &
  0.615 &
  0.567 &
  0.425 &
  0.642 &
  0.724 &
  0.964 &
  0.868 &
  0.927 &
  0.801 &
  0.977 &
  0.860 &
  0.761 \\
\multicolumn{14}{c}{\cellcolor{mygray}\textit{Average retention   rate $\le$ 11.1 \%}} \\
PDrop &
  0.373 &
  0.301 &
  0.248 &
  0.293 &
  0.177 &
  0.315 &
  0.263 &
  0.358 &
  0.406 &
  0.238 &
  0.102 &
  0.235 &
  0.276 \\
VisionZip &
  0.657 &
  0.520 &
  0.505 &
  oom &
  0.565 &
  0.661 &
  oom &
  0.660 &
  0.674 &
  oom &
  oom &
  oom &
  -- \\
VScan &
  0.635 &
  0.518 &
  0.475 &
  oom &
  0.484 &
  0.627 &
  oom &
  0.678 &
  0.781 &
  oom &
  oom &
  oom &
  -- \\
&
  \cellcolor{mygray}\textit{8.9\%} &
  \cellcolor{mygray}\textit{8.8\%} &
  \cellcolor{mygray}\textit{8.7\%} &
  \cellcolor{mygray}\textit{8.6\%} &
  \cellcolor{mygray}\textit{10.3\%} &
  \cellcolor{mygray}\textit{10.5\%} &
  \cellcolor{mygray}\textit{3.6\%} &
  \cellcolor{mygray}\textit{6.7\%} &
  \cellcolor{mygray}\textit{7.1\%} &
  \cellcolor{mygray}\textit{4.4\%} &
  \cellcolor{mygray}\textit{6.7\%} &
  \cellcolor{mygray}\textit{4.5\%} &
  \cellcolor{mygray}\textit{7.4\%} \\
\multirow{-2}{*}{GlimpsePrune} &
  \textbf{0.751} &
  \textbf{0.636} &
  \textbf{0.584} &
  \textbf{0.458} &
  \textbf{0.620} &
  \textbf{0.700} &
  \textbf{0.962} &
  \textbf{0.861} &
  \textbf{0.939} &
  \textbf{0.786} &
  \textbf{0.978} &
  \textbf{0.854} &
  \textbf{0.761} \\
\multicolumn{14}{c}{\cellcolor{mygray}\textit{Average retention   rate $\le$ 22.2 \%}} \\
PDrop &
  0.465 &
  0.380 &
  0.318 &
  0.354 &
  0.263 &
  0.486 &
  0.398 &
  0.473 &
  0.545 &
  0.352 &
  0.176 &
  0.288 &
  0.375 \\
VisionZip &
  0.716 &
  0.561 &
  0.534 &
  oom &
  0.598 &
  0.713 &
  oom &
  0.788 &
  0.840 &
  oom &
  oom &
  oom &
  -- \\
VScan &
  0.690 &
  0.568 &
  0.511 &
  oom &
  0.582 &
  0.645 &
  oom &
  0.775 &
  0.856 &
  oom &
  oom &
  oom &
  -- \\
 &
  \cellcolor{mygray}\textit{18.6\%} &
  \cellcolor{mygray}\textit{15.0\%} &
  \cellcolor{mygray}\textit{14.5\%} &
  \cellcolor{mygray}\textit{13.8\%} &
  \cellcolor{mygray}\textit{18.4\%} &
  \cellcolor{mygray}\textit{18.6\%} &
  \cellcolor{mygray}\textit{3.8\%} &
  \cellcolor{mygray}\textit{8.5\%} &
  \cellcolor{mygray}\textit{9.2\%} &
  \cellcolor{mygray}\textit{5.3\%} &
  \cellcolor{mygray}\textit{7.1\%} &
  \cellcolor{mygray}\textit{5.2\%} &
  \cellcolor{mygray}\textit{11.5\%} \\
\multirow{-2}{*}{GlimpsePrune} &
  \textbf{0.757} &
  \textbf{0.639} &
  \textbf{0.583} &
  \textbf{0.461} &
  \textbf{0.621} &
  \textbf{0.717} &
  \textbf{0.962} &
  \textbf{0.861} &
  \textbf{0.941} &
  \textbf{0.792} &
  \textbf{0.978} &
  \textbf{0.857} &
  \textbf{0.764} \\
\multicolumn{14}{c}{\cellcolor{mygray}\textit{Average retention   rate $\le$ 33.3 \%}} \\
PDrop &
  0.537 &
  0.427 &
  0.366 &
  0.392 &
  0.336 &
  0.552 &
  0.505 &
  0.562 &
  0.637 &
  0.420 &
  0.248 &
  0.392 &
  0.448 \\
VisionZip &
  0.731 &
  0.590 &
  0.550 &
  oom &
  0.619 &
  0.718 &
  oom &
  0.839 &
  0.897 &
  oom &
  oom &
  oom &
  --\\
VScan &
  0.714 &
  0.580 &
  0.515 &
  oom &
  0.590 &
  0.662 &
  oom &
  0.812 &
  0.892 &
  oom &
  oom &
  oom &
  --\\
 &
  \cellcolor{mygray}\textit{18.3\%} &
  \cellcolor{mygray}\textit{24.7\%} &
  \cellcolor{mygray}\textit{18.4\%} &
  \cellcolor{mygray}\textit{17.2\%} &
  \cellcolor{mygray}\textit{24.7\%} &
  \cellcolor{mygray}\textit{22.4\%} &
  \cellcolor{mygray}\textit{3.8\%} &
  \cellcolor{mygray}\textit{9.3\%} &
  \cellcolor{mygray}\textit{10.1\%} &
  \cellcolor{mygray}\textit{5.6\%} &
  \cellcolor{mygray}\textit{7.2\%} &
  \cellcolor{mygray}\textit{5.5\%} &
  \cellcolor{mygray}\textit{13.9\%} \\
\multirow{-2}{*}{GlimpsePrune} &
  \textbf{0.757} &
  \textbf{0.636} &
  \textbf{0.587} &
  \textbf{0.464} &
  \textbf{0.620} &
  \textbf{0.723} &
  \textbf{0.963} &
  \textbf{0.860} &
  \textbf{0.940} &
  \textbf{0.792} &
  \textbf{0.978} &
  \textbf{0.860} &
  \textbf{0.765} \\
\multicolumn{14}{c}{\cellcolor{mygray}\textit{Unlimited   constraint}} \\
 &
  \cellcolor{mygray}\textit{26.8\%} &
  \cellcolor{mygray}\textit{31.2\%} &
  \cellcolor{mygray}\textit{25.0\%} &
  \cellcolor{mygray}\textit{23.1\%} &
  \cellcolor{mygray}\textit{39.4\%} &
  \cellcolor{mygray}\textit{25.5\%} &
  \cellcolor{mygray}\textit{3.8\%} &
  \cellcolor{mygray}\textit{10.1\%} &
  \cellcolor{mygray}\textit{11.1\%} &
  \cellcolor{mygray}\textit{6.0\%} &
  \cellcolor{mygray}\textit{7.2\%} &
  \cellcolor{mygray}\textit{5.5\%} &
  \cellcolor{mygray}\textit{17.9\%} \\
\multirow{-2}{*}{GlimpsePrune} &
  0.759 &
  0.635 &
  0.583 &
  0.460 &
  0.618 &
  0.722 &
  \textbf{0.963} &
  0.860 &
  0.941 &
  0.795 &
  \textbf{0.978} &
  0.860 &
  0.764 \\
 &
  \cellcolor{mygray}\textit{28.7\%} &
  \cellcolor{mygray}\textit{31.9\%} &
  \cellcolor{mygray}\textit{24.9\%} &
  \cellcolor{mygray}\textit{23.2\%} &
  \cellcolor{mygray}\textit{45.9\%} &
  \cellcolor{mygray}\textit{23.7\%} &
  \cellcolor{mygray}\textit{6.9\%} &
  \cellcolor{mygray}\textit{12.1\%} &
  \cellcolor{mygray}\textit{11.5\%} &
  \cellcolor{mygray}\textit{9.9\%} &
  \cellcolor{mygray}\textit{13.8\%} &
  \cellcolor{mygray}\textit{8.4\%} &
  \cellcolor{mygray}\textit{20.1\%} \\
\multirow{-2}{*}{GlimpsePrune+} &
  \textbf{0.776} &
  \textbf{0.753} &
  \textbf{0.722} &
  \textbf{0.719} &
  \textbf{0.800} &
  \textbf{0.830} &
  0.956 &
  \textbf{0.876} &
  \textbf{0.953} &
  \textbf{0.822} &
  0.977 &
  \textbf{0.869} &
  \textbf{0.838} \\
\bottomrule
\end{tabular}
\caption{
  Performance in free-form VQA tasks, ``oom'' indicates out-of-memory errors.
}
\label{tab:free_form_vqa}
\end{table*}

\section{Experiments}

\subsection{Experimental setup}
\myPara{Model architectures.}
Our primary experiments are conducted on Qwen2.5-VL-7B~\cite{Qwen2.5-VL}, 
an open-source architecture that supports dynamically high-resolution inputs (from $4$ to $16,384$ visual tokens).
To demonstrate the generalizability of our approach, we also adapt and evaluate it on LLaVA-1.5-7B~\cite{liu2023improvedllava}, 
a model designed for fixed-resolution inputs ($576$ visual tokens). 
In these architectures, we set the VIP's hidden dimension to $E=256$ 
and the visual conditioning dimension to $F=512$,
with $M$ layers of visual features.
The pruning position is set to approximately $K=\lceil2/3 \times L\rceil$ of the original LLM decoder.
For detailed settings on different model sizes, please refer to the Appendix~\ref{app:arch_details}.

\begin{table*}[t]
\centering
\setlength{\tabcolsep}{4.5pt}
\small
\begin{tabular}{lccccccccccc}
\toprule
\textbf{Method} &
  \thead{\textsupsub{VQA}{v2}{vallite}} &
  \thead{GQA} &
  \thead{\textsub{VizWiz}{val}} &
  \thead{\textsup{SQA}{img}} &
  \thead{POPE} &
  \thead{MME} &
  \thead{\textsupsub{MMB}{en}{test}} &
  \thead{\textsupsub{MMB}{cn}{test}} &
  \thead{\textsup{SEED}{img}} &
  \thead{V*} &
  \thead{Average} \\
\midrule
Qwen2.5-VL-7B  & 79.5 & 61.4 & 70.6 & 87.5 & 87.5 & 2327.9 & 83.2 & 82.8 & 77.5 & 71.7 & 70.3 \\
\multicolumn{12}{c}{\cellcolor{mygray}\textit{Average retention   rate $\le$ 11.1 \%}} \\
PDrop         & 53.3 & 41.8 & 53.3 & 77.0 & 68.4 & 1642.2 & 42.9 & 38.7 & 49.4 & 42.4 & 46.8 \\
VisionZip     & 68.4 & 54.9 & 66.4 & 82.7 & 82.9 & 2031.6 & 75.6 & 72.6 &   --  & 60.2 &  --   \\
VScan         & 74.2 & 57.1 & 67.5 & 82.2 & 86.1 & 2129.8 & 74.1 & 71.5 &   --  & 59.7 &  --   \\
GlimpsePrune  & \textbf{78.0} & \textbf{59.5} & \textbf{69.4} & \textbf{87.2} & \textbf{87.5} & \textbf{2348.0} & \textbf{82.6} & \textbf{81.7} & \textbf{77.4} & \textbf{72.8} & \textbf{70.0} \\
\multicolumn{12}{c}{\cellcolor{mygray}\textit{Unlimited   constraint}}             \\
GlimpsePrune  & 79.1 & 60.0 & 70.5 & 87.3 & 87.6 & 2323.1 & 82.9 & 81.7 & 77.4 & 72.8 & 70.0 \\
\bottomrule
\end{tabular}
\caption{
  Performance in short-form VQA tasks, the maximum number of visual tokens is set to 2048 here.
}
\label{tab:short_form_vqa}
\end{table*}

\myPara{Training setups.}
To demonstrate the generalization capability of \Ours,
we train the VIP and glimpse token using only 20K randomly sampled data from the GQA dataset~\cite{hudson2019gqa}.
We only train for one epoch, which can be completed in about 0.5 hours on a single NVIDIA A100 GPU.
For the RL fine-tuning in \Ours+, we use 240K samples randomly selected from the VisCoT dataset~\cite{shao2024viscot},
which provides a diverse set of visual question-answering scenarios.
For the GRPO algorithm, we set the group size to 4, the number of policy update iteration to 1, 
and the KL divergence coefficient to 0.04, respectively.
We apply LoRA~\cite{hu2022lora} with a rank of $R=64$ to fine-tune the LLM decoder.
More details can be found in the Appendix~\ref{app:implementation_details}.

\begin{table*}[t]
\centering
\setlength{\tabcolsep}{1pt}
\small
\begin{tabular}{lccccccccccccc}
\toprule
 &
  \multicolumn{2}{c}{\thead{General \\ VQA}} &
  \multicolumn{3}{c}{\thead{Relation \\ Reasoning}} &
  \thead{Fine \\ grained} &
  \multicolumn{5}{c}{\thead{Doc/Text}} &
  \thead{Chart} &
   \\
\cmidrule(lr){2-3} \cmidrule(lr){4-6} \cmidrule(lr){7-7} \cmidrule(lr){8-12} \cmidrule(lr){13-13}
\multirow{-2}{*}{\thead{Method}} &
  Flickr30k &
  V7W &
  GQA &
  OpenImages &
  VSR &
  CUB &
  DocVQA &
  TextCaps &
  TextVQA &
  DUDE &
  SROIE &
  InfoVQA &
  \multirow{-2}{*}{\thead{Average}} \\
\midrule
LLaVA-1.5-7B &
  0.694 &
  0.658 &
  0.635 &
  0.519 &
  0.613 &
  0.560 &
  0.252 &
  0.632 &
  0.650 &
  0.293 &
  0.125 &
  0.393 &
  0.502 \\
\multicolumn{14}{c}{\cellcolor{mygray}\textit{Average retention rate $\le$ 11.1 \%}} \\
PDrop &
  0.493 &
  0.498 &
  0.454 &
  0.501 &
  0.462 &
  0.457 &
  0.134 &
  0.335 &
  0.314 &
  0.148 &
  0.022 &
  0.356 &
  0.348 \\
VisionZip &
  0.643 &
  0.577 &
  0.574 &
  0.511 &
  0.577 &
  0.502 &
  0.207 &
  0.595 &
  0.613 &
  0.242 &
  0.107 &
  0.349 &
  0.458 \\
DivPrune & 
  0.656 &
  0.616 &
  0.592 &
  0.522 &
  0.562 &
  0.483 &
  0.210 &
  0.533 &
  0.551 &
  0.230 &
  0.077 &
  \textbf{0.382} &
  0.451 \\
CDPruner &
  0.670 &
  0.616 &
  0.602 &
  0.531 &
  0.560 &
  0.506 &
  0.206 &
  0.551 &
  0.575 &
  0.217 &
  0.064 &
  0.368 &
  0.455 \\
VScan &
  0.663 &
  0.619 &
  0.598 &
  \textbf{0.540} &
  0.584 &
  0.516 &
  0.224 &
  \textbf{0.610} &
  \textbf{0.624} &
  0.244 &
  0.101 &
  0.370 &
  0.474 \\
 &
  \cellcolor{mygray}\textit{9.2\%} &
  \cellcolor{mygray}\textit{9.5\%} &
  \cellcolor{mygray}\textit{8.9\%} &
  \cellcolor{mygray}\textit{9.8\%} &
  \cellcolor{mygray}\textit{10.7\%} &
  \cellcolor{mygray}\textit{10.3\%} &
  \cellcolor{mygray}\textit{8.4\%} &
  \cellcolor{mygray}\textit{8.5\%} &
  \cellcolor{mygray}\textit{8.7\%} &
  \cellcolor{mygray}\textit{8.8\%} &
  \cellcolor{mygray}\textit{9.3\%} &
  \cellcolor{mygray}\textit{6.3\%} &
  \cellcolor{mygray}\textit{9.0\%} \\
\multirow{-2}{*}{GlimpsePrune} &
  \textbf{0.690} &
  \textbf{0.666} &
  \textbf{0.654} &
  0.526 &
  \textbf{0.615} &
  \textbf{0.547} &
  \textbf{0.236} &
  0.584 &
  0.582 &
  \textbf{0.273} &
  \textbf{0.124} &
  0.381 &
  \textbf{0.490} \\
\multicolumn{14}{c}{\cellcolor{mygray}\textit{Unlimited   constraint}} \\
 &
  \cellcolor{mygray}\textit{22.9\%} &
  \cellcolor{mygray}\textit{30.0\%} &
  \cellcolor{mygray}\textit{22.8\%} &
  \cellcolor{mygray}\textit{29.2\%} &
  \cellcolor{mygray}\textit{41.7\%} &
  \cellcolor{mygray}\textit{18.1\%} &
  \cellcolor{mygray}\textit{13.2\%} &
  \cellcolor{mygray}\textit{14.8\%} &
  \cellcolor{mygray}\textit{16.2\%} &
  \cellcolor{mygray}\textit{16.1\%} &
  \cellcolor{mygray}\textit{10.9\%} &
  \cellcolor{mygray}\textit{8.8\%} &
  \cellcolor{mygray}\textit{20.4\%} \\
\multirow{-2}{*}{GlimpsePrune} &
  0.693 &
  0.664 &
  0.649 &
  0.532 &
  0.606 &
  0.551 &
  0.247 &
  0.605 &
  0.621 &
  0.272 &
  0.119 &
  0.384 &
  0.495 \\
\bottomrule
\end{tabular}
\caption{Performance in free-form VQA tasks using LLaVA-1.5-7B.}
\label{tab:free_form_vqa_llava}
\end{table*}

\myPara{Evaluation setups.}
For free-form VQA, we utilize the comprehensive benchmark from VisCoT~\cite{shao2024viscot},
which comprises 12 diverse VQA datasets spanning multiple domains 
and featuring a wide range of target object scales. 
Crucially, evaluation is performed by a powerful LLM rather than rule-based string matching, 
providing a more holistic assessment of generative quality. 
We use Qwen2.5-32B-Instruct-GPTQ-Int8~\cite{Qwen2.5} as the evaluator LLM.
For short-form VQA, we select 10 widely-used VQA benchmarks 
and evaluate them under the LMMs-Eval framework~\cite{zhang2024lmmseval}.
These benchmarks typically require concise answers, such as a short phrase or a single option from a list, 
and are evaluated using exact-match or rule-based string comparison. 
Further details on these datasets are provided in the Appendix~\ref{app:dataset_details}.

\myPara{Comparison methods.}
We compare our method against several state-of-the-art and representative visual token compression techniques:
PDrop~\cite{xing2024pyramiddrop}, VisionZip~\cite{yang2025visionzip}, DivPrune~\cite{alvar2025divprune},
CDPruner~\cite{zhang2025cdpruner} and VScan~\cite{zhang2025vscan}.
Since these methods operate with a fixed compression ratio and apply compression at different stages,
we set an upper bound on the average retention rate 
of visual tokens' KV cache during the decoding phase for all methods.
As our proposed method, \Ours, features a dynamic retention rate, 
we will report both its performance under this constraint and its average retention rate on each dataset.

\begin{table*}[t]
\centering
\small
\begin{tabular}{ccccccccccc}
\toprule
\thead{Glimpse} & 
\bm{$\mathcal{L}_{lang}$} & 
\bm{$\mathcal{L}_{loc}$} & 
\thead{VIP} & 
\thead{Visual \\ Condition} &
\thead{General \\ VQA} &
\thead{Relation \\ Reasoning} &
\thead{Fine \\ Grained} &
\thead{Doc/Text} &
\thead{Chart} &
\thead{Average} \\
\midrule
\multicolumn{5}{c}{Qwen2.5-VL-3B}           & 0.664 & 0.531 & 0.748 & 0.892 & 0.816 & 0.746 \\
\midrule
\xmark & \xmark & \xmark & \xmark & \xmark & 0.585 & 0.515 & 0.762 & 0.506 & 0.542 & 0.546  \\
\cmark & \cmark & \xmark & \xmark & \xmark & 0.647 & 0.537 & 0.760 & 0.771 & 0.693 & 0.684  \\
\cmark & \cmark & \cmark & \xmark & \xmark & 0.641 & 0.521 & 0.763 & 0.802 & 0.802 & 0.702 \\
\cmark & \cmark & \cmark & \cmark & \xmark & 0.639 & 0.528 & 0.764 & 0.835 & 0.788 & 0.716 \\
\cmark & \cmark & \cmark & \cmark & \cmark & 0.642 & 0.524 & 0.760 & 0.882 & 0.804 & 0.736 \\
\bottomrule
\end{tabular}
\caption{Ablation study of training components and design choices in \Ours{}.}
\label{tab:ablation_glimpse}
\end{table*}

\begin{table}[ht]
\centering
\small
\begin{tabular}{ccc}
\toprule
\thead{Prune Layer~(K)} & \thead{Performance} & \thead{Prefilling FLOPs(T)} \\
\midrule
Qwen2.5-VL-3B             & 0.746                & 33.8                         \\
\midrule
18                       & 0.690                & 17.4                         \\
24                       & 0.736                & 23.1                         \\
30                       & 0.739                & 28.5                         \\
36                       & 0.556                & 34.1                         \\
\bottomrule
\end{tabular}
\caption{
  The trade-off between performance and efficiency with different pruning layers.
}
\label{tab:reduce_layer_analysis}
\end{table}

\subsection{Main results}

\myPara{\Ours{} for free-form VQA.}
As detailed in \tabref{tab:free_form_vqa}, we evaluate \Ours{} on 12 free-form VQA datasets 
under three distinct average retention rate constraints: 11.1\%, 22.2\%, and 33.3\%. 
With an average retention rate as low as 11.1\%, 
our method achieves a mean accuracy of 0.761, remarkably retaining 100\% of the performance of the original Qwen2.5-VL-7B model.

The dynamic nature of \Ours{} is particularly evident in its adaptability to varying target object scales. 
On datasets with small target objects, such as DocVQA, 
\Ours{} dynamically adjusts to a much lower retention rate (3.6-3.8\%) 
while exhibiting only a marginal performance drop (from 0.964 to 0.962). 
Conversely, on datasets featuring large-scale visual targets like VSR, 
even when constrained by a stringent upper bound,
\Ours{} effectively sparsifies the visual context. 
It reduces the retention rate from 39.4\% (unconstrained) to 10.3\% with similar performance (0.618 vs. 0.620),
These results strongly indicate that \Ours{} not only accurately identifies critical visual tokens 
but also intelligently sparsifies them when operating under tight computational budgets.
We provide qualitative examples in the Appendix~\ref{app:additional_results}.

On high-resolution inputs, methods like VisionZip~\cite{yang2025visionzip} and VScan~\cite{zhang2025vscan} face practical limitations. 
Their reliance on computing dense similarity matrices leads to prohibitive memory usage that negates FlashAttention2~\cite{dao2023flashattention2} benefits, 
often causing Out-of-Memory (OOM) errors. 
In contrast, our method is designed to remain efficient under these conditions.

\myPara{\Ours{} for short-form VQA.}
As shown in \tabref{tab:short_form_vqa},
\Ours{} also demonstrates strong performance on short-form VQA tasks.
With an average retention rate capped at 11.1\%,
\Ours{} achieves an average accuracy of 70.0\% across 10 benchmarks,
which is 99.6\% of the original Qwen2.5-VL-7B model's performance.

\myPara{\Ours{} for other architectures.}
As shown in \tabref{tab:free_form_vqa_llava},
\Ours{} also demonstrates strong performance on the LLaVA-1.5-7B model.
Due to the fixed number of 576 visual tokens used as input in LLaVA-1.5-7B,
the dynamic range for compression is smaller than that of the Qwen2.5-VL-7B model.
Nevertheless, \Ours{} still achieves an average accuracy of 0.490 with an average retention rate capped at 11.1\%,
which is 97.6\% of the original LLaVA-1.5-7B model's performance.
More results on other architectures can be found in the Appendix~\ref{app:additional_results}.

\begin{table}[ht]
\centering
\small
\begin{tabular}{lccc}
\toprule
\thead{Method} & \thead{Qwen2.5 \\ VL-7B} & \thead{Glimpse \\ Prune} & \thead{Glimpse \\ Prune+}  \\
\midrule
Prefill FLOPs (T)                      & 77.8         & 53.8         & 54.2  \\
Prefill Time (ms/token)                & 423.3        & 306.9        & 309.9  \\
Length of KV Cache                     & 5073.9       & 202.5        & 291.1 \\
Decode FLOPs (B)                       & 333.4        & 317.4        & 256.6 \\
Decode Time (ms/token)                 & 29.5         & 28.3         & 28.4  \\
\#Generate tokens                      & 22.0         & 24.2         & 19.5 \\
Max GPU Memory (GB)                    & 33.5         & 24.4         & 23.7  \\
Average Score                          & 0.761        & 0.761        & 0.835 \\
\bottomrule
\end{tabular}
\caption{
Efficiency analysis of \Ours{} on Qwen2.5-VL-7B model with 100 samples from DocVQA.
}
\label{tab:efficiency}
\end{table}

\begin{table}[h]
\centering
\small
\begin{tabular}{cccc}
\toprule
\thead{Pruned} & \thead{Time (s/iter)} &  \thead{Memory (GB)} & \thead{Performance} \\
\midrule
\xmark & 37.1 & $\ge 80 \times 4$   & 0.841  \\
\cmark & 29.9 & $\sim 56 \times 4$  & 0.835 \\
\bottomrule
\end{tabular}
\caption{
\Ours{} reduces the cost of RL fine-tuning.
Experiments are conducted on 4 NVIDIA A100 GPUs.
}
\label{tab:efficiency_rl}
\end{table}

\subsection{Ablation studies}\label{sec:ablations}
We conduct ablation studies on the Qwen2.5-VL-3B model,
which has an LLM decoder with $L=36$ layers and achieves an average performance of 0.746 on free-form VQA tasks.
We will analyze the impact of design choices and training components in \Ours{} under a fixed retention rate upper limit of 11.1\%.

\myPara{The importance of glimpse tokens.}
We first implement a baseline without training, 
as shown in the first row of \tabref{tab:ablation_glimpse}:
using the attention weights from the first generated token to the visual tokens at layer $K=24$ as the pruning metric.
This baseline achieves an average accuracy of only 0.546.
When we add glimpse tokens and subsequently add language and location losses,
the performance gradually improves to 0.684 and 0.702, respectively.
This demonstrates that glimpse tokens can provide a good visual importance prediction before the model generates free-form answers.

\myPara{The importance of visual conditions.}
The last two rows of \tabref{tab:ablation_glimpse} highlight the necessity of 
conditioning our importance map optimization on visual features. 
Without them, merely adding a 4-layer self-attention VIP only improves the performance from 0.702 to 0.716.
Conversely, incorporating visual features as conditions in VIP boosts performance to 0.736. 
Notably, this performance gain predominantly stems from Doc/Text data, 
whose distribution significantly differs from the training set. 
This suggests that visual conditioning mitigates the model's tendency towards shortcut learning.

\myPara{The selection of pruning layer.}
The choice of the pruning layer, $K$, 
presents a trade-off between performance and efficiency. 
While deeper layers provide more interpretable attention maps, 
they also increase prefill computation. 
Based on our analysis in \tabref{tab:reduce_layer_analysis}, 
we found $K=24$ (two-thirds of the total layers) to be an optimal balance. 
We therefore adopt this $2/3$ ratio across all tested architectures.
Notably, pruning at $K=36$ (the last layer) leads to a performance drop to 0.556,
as the glimpse attention of the final layer tends to focus less on visual information.

\subsection{Efficiency analysis}

\myPara{Efficent inference.}
As shown in \tabref{tab:efficiency}, 
our method significantly enhances inference efficiency. 
Benchmarked on a single A100 GPU with the Qwen2.5-VL-7B model on 100 DocVQA samples (with KV Cache and FlashAttention2), 
\Ours{} reduces the computation-intensive prefill cost to 69.1\% of the baseline. 
More critically, for the memory-intensive decoding phase, it slashes the initial KV Cache length from 5,073.9 to 202.5 tokens. 
This leads to a reduction in peak memory usage across the entire generation process to 72.8\%.

\myPara{Efficient RL fine-tuning.}
Our fine-tuning experiments on the Qwen2.5-VL-7B model (\tabref{tab:efficiency_rl}) demonstrate that our pruning method effectively addresses Out-of-Memory (OOM) errors encountered when training with token lengths over 3000.
By applying this technique, we not only supported token length to 6000 but also reduced the average iteration time and GPU memory usage to 81\% and 70\% of the original, respectively. 
Crucially, this efficiency gain was achieved with negligible impact on performance (0.835 vs. 0.841 compared to the dense-token baseline).

\section{Conclusions}

This work demonstrates the failure of static token compression methods and introduces \Ours{}, 
a dynamic visual token pruning framework. By learning a data-driven pruning metric, 
\Ours{} prunes 92.6\% of visual tokens while on average retaining 100\% of the baseline performance.
An enhanced version, \Ours{}+, further boosts performance to 110\% using GRPO fine-tuning. 
Our method presents a practical and robust solution for building powerful and efficient LVLMs.

{
\small
\bibliographystyle{unsrt}
\bibliography{arxiv}
}

\clearpage
\appendix

\section{Dataset Details}\label{app:dataset_details}

\subsection{Free-form VQA datasets}\label{app:free_form_datasets}
The free-form VQA tasks are evaluated on the VisCoT benchmark~\cite{shao2024viscot}, 
which consists of 12 datasets.
\tabref{tab:free_form_datasets} lists the number of samples 
and the average ratio of target objects in each dataset.

\begin{table}[ht]
\centering
\small
\setlength{\tabcolsep}{3pt}
\begin{tabular}{lcrlcr}
\toprule
\thead{Dataset} & \thead{\#Samples} & \thead{Ratio} & \thead{Dataset} & \thead{\#Samples} & \thead{Ratio} \\
\midrule
Flickr30k       & 1546              & 18.2\%                   & DocVQA          & 888               & 6.8\%                    \\
V7W             & 1000              & 22.8\%                   & TextCaps        & 853               & 4.9\%                    \\
GQA             & 978               & 20.2\%                   & TextVQA         & 526               & 5.6\%                    \\
OpenImages      & 945               & 24.3\%                   & DUDE            & 603               & 1.7\%                    \\
VSR             & 404               & 41.6\%                   & SROIE           & 686               & 10.1\%                   \\
CUB             & 492               & 42.5\%                   & InfoVQA         & 360               & 12.3\%                   \\
\bottomrule
\end{tabular}
\caption{
  Datasets used for free-form VQA evaluation.
  ``Ratio'' refers to the average ratio of target objects in the images.
}
\label{tab:free_form_datasets}
\end{table}

\subsection{Short-form VQA datasets}\label{app:short_form_datasets}
\tabref{tab:short_form_datasets} lists the number of samples and answer types for the datasets used in short-form VQA evaluation,
including VQAv2~\cite{goyal2017vqav2}, GQA~\cite{hudson2019gqa}, VizWiz~\cite{gurari2018vizwiz}, ScienceQA~\cite{lu2022sqa},
POPE~\cite{li2023pope}, MME~\cite{fu2024mme}, MMBench~\cite{liu2024mmbench}, SEEDBench~\cite{li2023seed}, and V* bench~\cite{wu2024vstar}.

\begin{table}[ht]
\centering
\small
\setlength{\tabcolsep}{3pt}
\begin{tabular}{lcrlcr}
\toprule
\thead{Dataset}               & \thead{\#Samples} & \thead{Type} & \thead{Dataset}            & \thead{\#Samples} & \thead{Type} \\
\midrule
\textsupsub{VQA}{v2}{vallite} & 500               & Phrase       & MME                        & 2374              & Phrase    \\
GQA                           & 12578             & Phrase       & \textsupsub{MMB}{cn}{test} & 6666              & Selection \\
\textsub{VizWiz}{val}         & 4319              & Phrase       & \textsupsub{MMB}{en}{test} & 6666              & Selection \\
\textsup{SQA}{img}            & 2017              & Selection    & \textsup{SEED}{img}        & 17990             & Selection \\
POPE                          & 9000              & Judgment     & V*                         & 191               & Selection \\
\bottomrule
\end{tabular}
\caption{
  Datasets used for short-form VQA evaluation.
  The ``Type'' column indicates the type of answers in the dataset.
  ``Phrase'' means the answer is a word or phrase, 
  ``Selection'' means the answer is selected from a set of options, 
  and ``Judgment'' means the answer is a judgment (e.g., yes/no).
}
\label{tab:short_form_datasets}
\end{table}

\subsection{Datasets for training \Ours}\label{app:train_datasets}
We train \Ours{} using 20,000 randomly sampled samples from the GQA dataset~\cite{hudson2019gqa}.
Each sample contains an image, a question, a short answer, and several bounding boxes of visual targets.
We find that \Ours{} generalizes well to various VQA datasets after training on GQA,
allowing our method to operate without the need for further expansion of the training dataset.

\subsection{Datasets for RL fine-tuning}\label{app:rl_datasets}
Since fine-tuning datasets for LLMs often require domain richness,
we use the training sets of 12 datasets from VisCoT~\cite{shao2024viscot} for fine-tuning the LLM.
Specifically, we randomly sample 20,000 samples from each dataset,
with 10,000 samples constructed as short-form VQA tasks and 10,000 samples as free-form VQA tasks.
Due to the high image resolution or long answers in some data, which can easily lead to out-of-memory error during training,
we limit the number of input tokens to 6000, and only compute the GRPO loss~\cite{shao2024deepseekmath} for the first 128 generated tokens.

\section{Implementation Details}\label{app:implementation_details}
\subsection{Training details.}\label{app:train_details}
\tabref{tab:train_details} lists the hyperparameters used for training \Ours{},
and \tabref{tab:train_rl_details} lists the hyperparameters used for RL fine-tuning.

\begin{table}[ht]
\centering
\small
\begin{tabular}{lrlr} 
\toprule
\thead{Key}      & \thead{Value} & \thead{Key}      & \thead{Value} \\
\midrule
batch / device              & 1              & dtype                                     & bfloat16       \\
grad acc steps              & 8              & max grad norm                           & 1.0            \\
num gpus                    & 2              & train data size                         & 20,000          \\
learning rate               & 1e-4           & num train epochs                        & 1              \\
warmup ratio                & 0.1            & weight of $\mathcal{L}_\text{language}$ & 1.0            \\
optimizer                   & AdamW          & weight of $\mathcal{L}_\text{dice}$     & 1.0            \\
scheduler                   & cosine         & weight of $\mathcal{L}_\text{bce}$      & 0.1            \\
\bottomrule
\end{tabular}
\caption{
Training arguments for \Ours{}.
}
\label{tab:train_details}
\end{table}

\begin{table}[ht]
\centering
\small
\begin{tabular}{lrlr}
\toprule
\thead{Key}      & \thead{Value} & \thead{Key} & \thead{Value} \\
\midrule
batch / device          & 1         & optimizer              & AdamW \\
grad acc steps          & 8         & scheduler              & cosine \\
num gpus                & 4         & warmup ratio           & 0.1 \\
num generations         & 4         & dtype                  & bfloat16 \\
num iterations          & 1         & max grad norm          & 1.0 \\
lora rank               & 32        & train data size        & 240,000 \\
lora alpha              & 64        & num train epochs       & 1 \\
lora dropout            & 0.05      & weight of $\mathcal{L}_\text{policy}$ & 1.0 \\
learning rate           & 1e-4      & weight of $\mathcal{L}_\text{KL}$ & 0.04 \\
reward model            & \multicolumn{3}{c}{Qwen2.5-32B-Instruct-GPTQ-Int8} \\
\bottomrule
\end{tabular}
\caption{
Training arguments for RL fine-tuning.
}
\label{tab:train_rl_details}
\end{table}

\begin{table}[ht]
\centering
\small
\setlength{\tabcolsep}{3pt}
\begin{tabular}{lrlr}
\toprule
\multicolumn{2}{c}{\thead{VQA  LVLM}} & \multicolumn{2}{c}{\thead{Evaluator LLM}} \\
\cmidrule(lr){1-2} \cmidrule(lr){3-4}
temperature              & 0.0             & temperature        & 0.7                            \\
dtype                    & bfloat16        & top p              & 0.8                            \\
use cache                & True            & repetition penalty & 1.05                           \\
max new tokens           & 1024            & max new tokens     & 512                            \\
attn impl                & FlashAttention2 &                    &                               \\
\bottomrule
\end{tabular}
\caption{
Evaluation details for free-form VQA tasks.
}
\label{tab:eval_details}
\end{table}

\begin{figure*}[ht]
    \centering
    \begin{mybox}[0.9\linewidth]
    You are responsible for proofreading the answers, you need to give a score to the model's answer by referring to the standard answer, based on the given question. The full score is 1 point and the minimum score is 0 points. Please output the score in the form ``score: $<$score$>$''. The evaluation criteria require that the closer the model's answer is to the standard answer, the higher the score. \\
    question: \texttt{\{\}} \\
    standard answer: \texttt{\{\}} \\
    model's answer: \texttt{\{\}} \\
    \end{mybox}
    \vspace{-1em}
    \caption{
      Input template for the evaluator LLM.
    }
    \label{fig:evaluate_template}
\end{figure*}

\begin{table*}[htb]
\centering
\small
\setlength{\tabcolsep}{10pt}
\begin{tabular}{lcccc}
\toprule
\thead{Architecture}        & \thead{Qwen2.5VL-3B}    & \thead{Qwen2.5VL-7B}    & \thead{LLaVA-1.5-7B}  & \thead{LLaVA-1.5-13B} \\
\midrule
Num visual tokens             & 4$\sim$16384            & 4$\sim$16384            & 576                   & 576  \\
Num visual layers             & 32                      & 32                      & 24                    & 24   \\
Visual feature size ($C$)   & 1280                    & 1280                    & 1024                  & 1024 \\
Num LLM layers ($L$)          & 36                      & 28                      & 32                    & 40   \\
Num LLM attention heads ($H$) & 16                      & 28                      & 32                    & 40   \\
LLM hidden size ($D$)       & 2048                    & 3584                    & 4096                  & 5120 \\
\midrule
Selected visual layers      & [31, 23, 15, 7]         & [31, 23, 15, 7]         & [23, 17, 11, 5]       & [23, 17, 11, 5]           \\
Glimpse token shape         & 36 $\times$ 2048        & 28 $\times$ 3584        & 32 $\times$ 4096      & 40 $\times$ 5120      \\
Prune Layer ($K$)           & 24                      & 19                      & 22                    & 27   \\
VIP hidden size ($E$)       & 256                     & 256                     & 256                   & 256  \\
VIP condition size ($F$)    & 512                     & 512                     & 512                   & 512  \\
VIP attention heads         & 4                       & 4                       & 4                     & 4    \\                  
\bottomrule
\end{tabular}
\caption{
Details of the model architectures.
}
\label{tab:arch_details}
\end{table*}

\subsection{Evaluation details.}\label{app:evaluation_details}
For free-form VQA tasks, we use Qwen2.5-32B-Instruct-GPTQ-Int8 as the evaluator LLM.
The evaluation template is shown in \figref{fig:evaluate_template}.
The evaluation settings for the LVLMs and the evaluator LLM are shown in \tabref{tab:eval_details}.

For short-form VQA tasks, we use the default configuration of the LMMs-Eval framework~\cite{zhang2024lmmseval}.

\section{Model Architecture Details}\label{app:arch_details}

As demonstrated in the main paper, 
we validated the effectiveness of our method, \Ours{}, on several baseline models, 
including Qwen2.5-VL~\cite{Qwen2.5-VL} and LLaVA-1.5~\cite{liu2023improvedllava}. 
The detailed architectural parameters for these models are presented in \tabref{tab:arch_details}.

Our approach introduces two primary new components: 
learnable glimpse tokens and a VIP module composed of $M$ self-attention blocks. 
Specifically, the glimpse tokens are integrated into every layer of the LLM decoder. 
However, during inference, only the tokens in the first $K$ layers are utilized. 
The position of pruning, $K$, is determined based on our ablation studies (\secref{sec:ablations}), 
following the rule $K=\lceil 2/3 \times L \rceil$, where $L$ is the total number of decoder layers. 
Furthermore, the VIP module is conditioned on $M=4$ visual features that are uniformly sampled from the visual encoder.

\section{Additional Results and Analysis}\label{app:additional_results}

\subsection{Results on other architectures.}\label{app:additional_results_arch}
In addition to the results presented in \tabref{tab:free_form_vqa} and~\tabref{tab:free_form_vqa_llava} of the main paper,
\tabref{tab:free_form_vqa_qwen_3b} and~\tabref{tab:free_form_vqa_llava_13b} 
present the performance of \Ours{} on the free-form VQA task, based on the Qwen2.5-VL-3B and LLaVA-1.5-13B models, respectively.

Compared to previous methods, our approach demonstrates a superior ability to dynamically adjust the retention rate 
while achieving better performance. 
This highlights the adaptability and efficiency of \Ours{} across different model scales and architectures.

\begin{table*}[ht]
\centering
\setlength{\tabcolsep}{1pt}
\small
\begin{tabular}{lccccccccccccc}
\toprule
 &
  \multicolumn{2}{c}{\thead{General \\ VQA}} &
  \multicolumn{3}{c}{\thead{Relation \\ Reasoning}} &
  \thead{Fine \\ grained} &
  \multicolumn{5}{c}{\thead{Doc/Text}} &
  \thead{Chart} &
   \\
\cmidrule(lr){2-3} \cmidrule(lr){4-6} \cmidrule(lr){7-7} \cmidrule(lr){8-12} \cmidrule(lr){13-13}
\multirow{-2}{*}{\thead{Method}} &
  Flickr30k &
  V7W &
  GQA &
  OpenImages &
  VSR &
  CUB &
  DocVQA &
  TextCaps &
  TextVQA &
  DUDE &
  SROIE &
  InfoVQA &
  \multirow{-2}{*}{\thead{Average}} \\
\midrule
Qwen2.5VL-3B &
  0.703 &
  0.626 &
  0.584 &
  0.438 &
  0.573 &
  0.748 &
  0.944 &
  0.840 &
  0.927 &
  0.793 &
  0.957 &
  0.816 &
  0.746 \\
\multicolumn{14}{c}{\cellcolor{mygray}\textit{Average retention rate $\le$ 11.1\%}} \\
PDrop &
  0.401 &
  0.298 &
  0.267 &
  0.378 &
  0.222 &
  0.502 &
  0.254 &
  0.428 &
  0.463 &
  0.255 &
  0.125 &
  0.320 &
  0.326 \\
VisionZip &
  0.546 &
  0.478 &
  0.465 &
  oom &
  0.426 &
  0.713 &
  oom &
  0.600 &
  0.603 &
  oom &
  oom &
  oom &
  -- \\
VScan &
  0.492 &
  0.444 &
  0.402 &
  oom &
  0.367 &
  0.717 &
  oom &
  0.564 &
  0.618 &
  oom &
  oom &
  oom &
  -- \\
 &
  \cellcolor{mygray}\textit{8.9\%} &
  \cellcolor{mygray}\textit{8.9\%} &
  \cellcolor{mygray}\textit{8.8\%} &
  \cellcolor{mygray}\textit{9.2\%} &
  \cellcolor{mygray}\textit{10.6\%} &
  \cellcolor{mygray}\textit{10.4\%} &
  \cellcolor{mygray}\textit{5.2\%} &
  \cellcolor{mygray}\textit{7.6\%} &
  \cellcolor{mygray}\textit{7.7\%} &
  \cellcolor{mygray}\textit{7.1\%} &
  \cellcolor{mygray}\textit{7.3\%} &
  \cellcolor{mygray}\textit{6.9\%} &
  \cellcolor{mygray}\textit{8.2\%} \\
\multirow{-2}{*}{GlimpsePrune} &
  \textbf{0.675} &
  \textbf{0.609} &
  \textbf{0.567} &
  \textbf{0.444} &
  \textbf{0.561} &
  \textbf{0.760} &
  \textbf{0.934} &
  \textbf{0.830} &
  \textbf{0.907} &
  \textbf{0.774} &
  \textbf{0.964} &
  \textbf{0.804} &
  \textbf{0.736} \\
\multicolumn{14}{c}{\cellcolor{mygray}\textit{Unlimited   constraint}} \\
 &
  \cellcolor{mygray}\textit{25.4\%} &
  \cellcolor{mygray}\textit{31.8\%} &
  \cellcolor{mygray}\textit{25.9\%} &
  \cellcolor{mygray}\textit{27.0\%} &
  \cellcolor{mygray}\textit{44.1\%} &
  \cellcolor{mygray}\textit{23.0\%} &
  \cellcolor{mygray}\textit{4.9\%} &
  \cellcolor{mygray}\textit{11.7\%} &
  \cellcolor{mygray}\textit{13.3\%} &
  \cellcolor{mygray}\textit{9.3\%} &
  \cellcolor{mygray}\textit{6.5\%} &
  \cellcolor{mygray}\textit{7.3\%} &
  \cellcolor{mygray}\textit{19.2\%} \\
\multirow{-2}{*}{GlimpsePrune} &
  0.695 &
  0.624 &
  0.578 &
  0.437 &
  0.560 &
  0.767 &
  0.931 &
  0.828 &
  0.916 &
  0.776 &
  0.965 &
  0.813 &
  0.741 \\
\bottomrule
\end{tabular}
\caption{Performance in free-form VQA tasks using Qwen2.5VL-3B.}
\label{tab:free_form_vqa_qwen_3b}
\end{table*}

\begin{table*}[ht]
\centering
\setlength{\tabcolsep}{1pt}
\small
\begin{tabular}{lccccccccccccc}
\toprule
 &
  \multicolumn{2}{c}{\thead{General \\ VQA}} &
  \multicolumn{3}{c}{\thead{Relation \\ Reasoning}} &
  \thead{Fine \\ grained} &
  \multicolumn{5}{c}{\thead{Doc/Text}} &
  \thead{Chart} &
   \\
\cmidrule(lr){2-3} \cmidrule(lr){4-6} \cmidrule(lr){7-7} \cmidrule(lr){8-12} \cmidrule(lr){13-13}
\multirow{-2}{*}{\thead{Method}} &
  Flickr30k &
  V7W &
  GQA &
  OpenImages &
  VSR &
  CUB &
  DocVQA &
  TextCaps &
  TextVQA &
  DUDE &
  SROIE &
  InfoVQA &
  \multirow{-2}{*}{\thead{Average}} \\
\midrule
LLaVA-1.5-13B &
  0.719 &
  0.674 &
  0.656 &
  0.533 &
  0.643 &
  0.608 &
  0.273 &
  0.660 &
  0.674 &
  0.301 &
  0.153 &
  0.407 &
  0.525 \\
\multicolumn{14}{c}{\cellcolor[HTML]{E7E6E6}\textit{Average retention rate $\le$ 11.1\%}} \\
PDrop &
  0.604 &
  0.581 &
  0.564 &
  0.556 &
  0.551 &
  0.539 &
  0.163 &
  0.477 &
  0.440 &
  0.173 &
  0.032 &
  0.402 &
  0.423 \\
VisionZip &
  0.662 &
  0.593 &
  0.581 &
  0.517 &
  0.580 &
  0.602 &
  0.240 &
  0.622 &
  0.633 &
  0.243 &
  0.109 &
  0.387 &
  0.481 \\
DivPrune &
  0.662 &
  0.620 &
  0.601 &
  0.513 &
  0.582 &
  0.562 &
  0.219 &
  0.543 &
  0.586 &
  0.236 &
  0.071 &
  0.394 &
  0.466 \\
CDPruner &
  \textbf{0.682} &
  0.639 &
  0.609 &
  0.514 &
  0.585 &
  \textbf{0.630} &
  0.218 &
  0.583 &
  0.602 &
  0.244 &
  0.069 &
  0.381 &
  0.480 \\
VScan &
  0.673 &
  0.625 &
  0.620 &
  \textbf{0.530} &
  0.600 &
  0.604 &
  0.247 &
  0.625 &
  0.643 &
  0.267 &
  0.111 &
  0.397 &
  0.495 \\
 &
  \cellcolor{mygray}\textit{9.5\%} &
  \cellcolor{mygray}\textit{9.5\%} &
  \cellcolor{mygray}\textit{9.0\%} &
  \cellcolor{mygray}\textit{10.0\%} &
  \cellcolor{mygray}\textit{10.6\%} &
  \cellcolor{mygray}\textit{9.4\%} &
  \cellcolor{mygray}\textit{10.2\%} &
  \cellcolor{mygray}\textit{9.3\%} &
  \cellcolor{mygray}\textit{9.4\%} &
  \cellcolor{mygray}\textit{10.6\%} &
  \cellcolor{mygray}\textit{10.0\%} &
  \cellcolor{mygray}\textit{9.7\%} &
  \cellcolor{mygray}\textit{9.8\%} \\
\multirow{-2}{*}{GlimpsePrune} &
  0.677 &
  \textbf{0.691} &
  \textbf{0.673} &
  0.523 &
  \textbf{0.623} &
  0.595 &
  \textbf{0.273} &
  \textbf{0.631} &
  \textbf{0.655} &
  \textbf{0.283} &
  \textbf{0.148} &
  \textbf{0.418} &
  \textbf{0.516} \\
\multicolumn{14}{c}{\cellcolor[HTML]{E7E6E6}\textit{Unlimited   constraint}} \\
 &
  \cellcolor{mygray}\textit{27.0\%} &
  \cellcolor{mygray}\textit{32.8\%} &
  \cellcolor{mygray}\textit{23.8\%} &
  \cellcolor{mygray}\textit{31.9\%} &
  \cellcolor{mygray}\textit{42.4\%} &
  \cellcolor{mygray}\textit{14.3\%} &
  \cellcolor{mygray}\textit{21.8\%} &
  \cellcolor{mygray}\textit{17.2\%} &
  \cellcolor{mygray}\textit{17.4\%} &
  \cellcolor{mygray}\textit{28.9\%} &
  \cellcolor{mygray}\textit{17.4\%} &
  \cellcolor{mygray}\textit{26.0\%} &
  \cellcolor{mygray}\textit{25.1\%} \\
\multirow{-2}{*}{GlimpsePrune} &
  0.687 &
  0.684 &
  0.673 &
  0.521 &
  0.625 &
  0.605 &
  0.273 &
  0.636 &
  0.663 &
  0.302 &
  0.156 &
  0.423 &
  0.521 \\
\bottomrule
\end{tabular}
\caption{
Performance in free-form VQA tasks using LLaVA-1.5-13B.
}
\label{tab:free_form_vqa_llava_13b}
\end{table*}

\subsection{Case studies.}\label{app:case_studies}
\begin{figure*}[ht]
    \begin{fullwidthsidebysidebox}[righthand width=0.65\textwidth]{myredbg}
        \includegraphics[width=\linewidth]{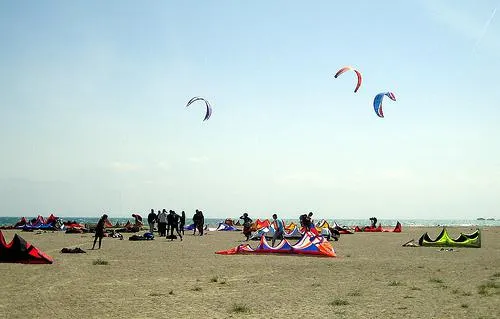}
        \tcblower
        \textbf{Question:} What is in the sky? \\
        \textbf{Standard answer:} Kites
        \par\medskip
        \textbf{Qwen2.5-VL-7B:} In the sky, there are \yellowhighlight{several kites}. The kites appear to be of different shapes and colors, and they are flying at varying heights. The sky is clear, suggesting a good day for kite flying.
        \par\medskip
        \textbf{Score:} 0.8
    \end{fullwidthsidebysidebox}

    \begin{fullwidthsidebysidebox}[righthand width=0.65\textwidth]{mypurplebg}
        \includegraphics[width=\linewidth]{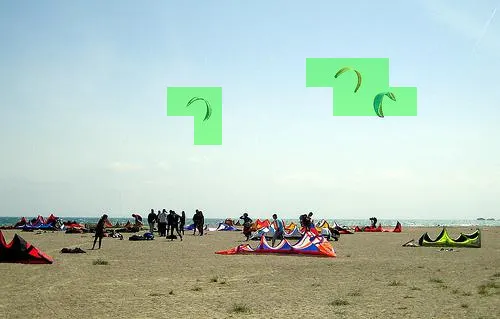}
        \tcblower
        \textbf{Retention ratio:} 4.5\%
        \par\medskip
        \textbf{\Ours{}:} In the sky, there are kites. There are \yellowhighlight{at least three kites} visible, each with a distinct shape and color.
        \par\medskip
        \textbf{Score:} 1.0
    \end{fullwidthsidebysidebox}
    \caption{Case 1:
    With only the relevant visual tokens retained, \Ours{} can focus more on the target object rather than other irrelevant information while answering the question correctly.
    }
    \label{fig:case_1}
\end{figure*}

\begin{figure*}[ht]
    \begin{fullwidthsidebysidebox}[righthand width=0.7\textwidth]{myredbg}
        \includegraphics[width=\linewidth]{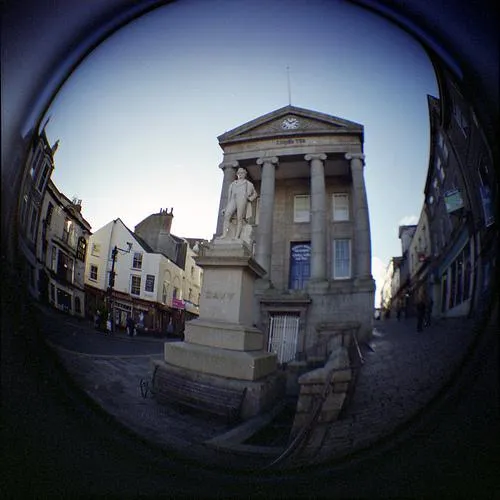}
        \tcblower 
        \textbf{Question:} What is on top of the building? \\
        \textbf{Standard answer:} A clock.
        \par\medskip
        \textbf{Qwen2.5-VL-7B:} The image shows a building \redhighlight{with a statue on top of it}. The statue appears to be of a person, possibly a historical figure, given its prominent placement and the formal setting. The building itself has a classical architectural style, with columns and a pediment, which is typical of neoclassical architecture.
        \par\medskip
        \textbf{Score:} 0.0
    \end{fullwidthsidebysidebox}

    \begin{fullwidthsidebysidebox}[righthand width=0.7\textwidth]{mypurplebg}
        \includegraphics[width=\linewidth]{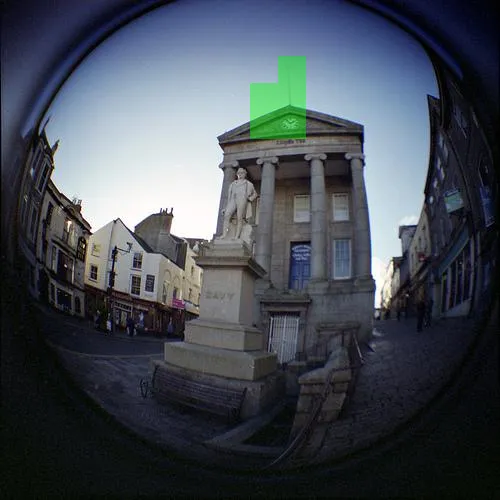}
        \tcblower 
        \textbf{Retention ratio:} 1.5\%
        \par\medskip
        \textbf{\Ours{}:} The image shows a building with a triangular pediment on top. On the pediment, \yellowhighlight{there is a clock}. The clock appears to be a decorative element, likely used to tell time for the building's occupants or visitors. The pediment itself is a classical architectural feature.
        \par\medskip
        \textbf{Score:} 0.7
    \end{fullwidthsidebysidebox}
    \caption{Case 2:
    By retaining the visual tokens at the correct position, \Ours{} can accurately answer that there is a clock on top of the building and provide relevant descriptions.
    }
    \label{fig:case_2}
\end{figure*}

\begin{figure*}[ht]
    \begin{fullwidthsidebysidebox}[righthand width=0.7\textwidth]{myredbg}
        \includegraphics[width=\linewidth]{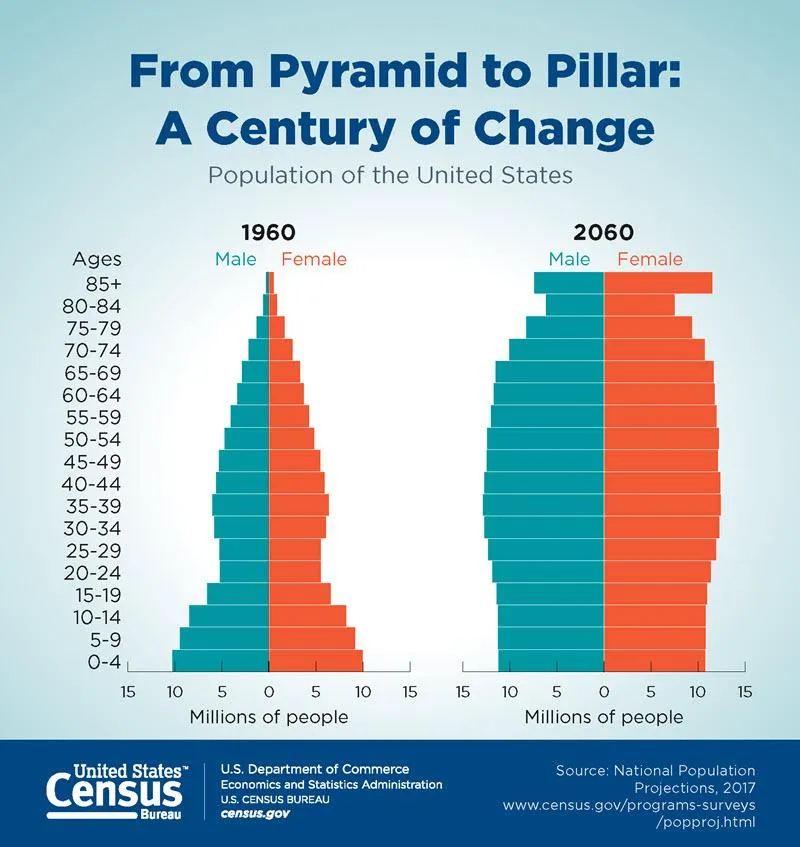}
        \tcblower
        \textbf{Question:} In 1960, what is the population of females in the age group 0-4 (millions)?\\
        \textbf{Standard answer:} 10
        \par\medskip
        \textbf{Qwen2.5-VL-7B:} In 1960, the population of females in the age group 0-4 is approximately \redhighlight{5} million.
        \par\medskip
        \textbf{Score:} 0.5
    \end{fullwidthsidebysidebox}

    \begin{fullwidthsidebysidebox}[righthand width=0.7\textwidth]{mypurplebg}
        \includegraphics[width=\linewidth]{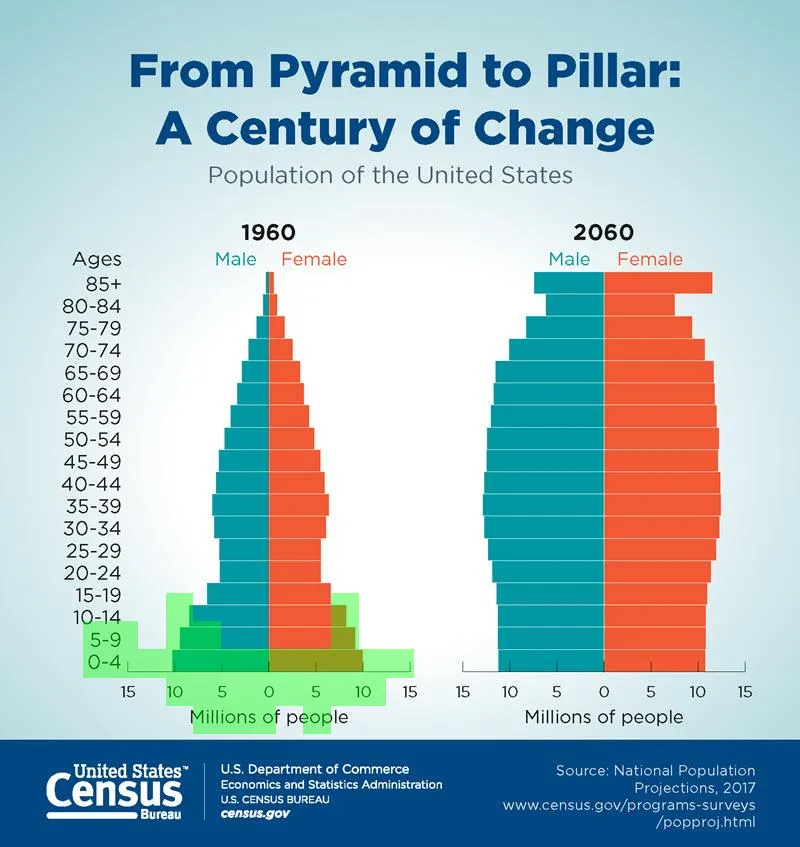}
        \tcblower
        \textbf{Retention ratio:} 3.6\%
        \par\medskip
        \textbf{\Ours{}:} In 1960, the population of females in the age group 0-4 is approximately \yellowhighlight{10} million.
        \par\medskip
        \textbf{Score:} 1.0
    \end{fullwidthsidebysidebox}
    \caption{Case 3:
    With the correct visual tokens retained, \Ours{} can answer the question related to the chart.
    }
    \label{fig:case_3}
\end{figure*}

\begin{figure*}[ht]
    \begin{fullwidthsidebysidebox}[righthand width=0.7\textwidth]{myredbg}
        \includegraphics[width=\linewidth]{figures/cases/failure/docvqa_708}
        \tcblower
        \textbf{Question:} What has been the greatest concern for non-smokers in 1990?\\
        \textbf{Standard answer:} Getting enough exercise
        \par\medskip
        \textbf{Qwen2.5-VL-7B:} The greatest concern for non-smokers in 1990, according to the table, is ``\yellowhighlight{Getting enough exercise},'' with 45\% of non-smokers being concerned about it.
        \par\medskip
        \textbf{Score:} 1.0
    \end{fullwidthsidebysidebox}

    \begin{fullwidthsidebysidebox}[righthand width=0.7\textwidth]{mypurplebg}
        \includegraphics[width=\linewidth]{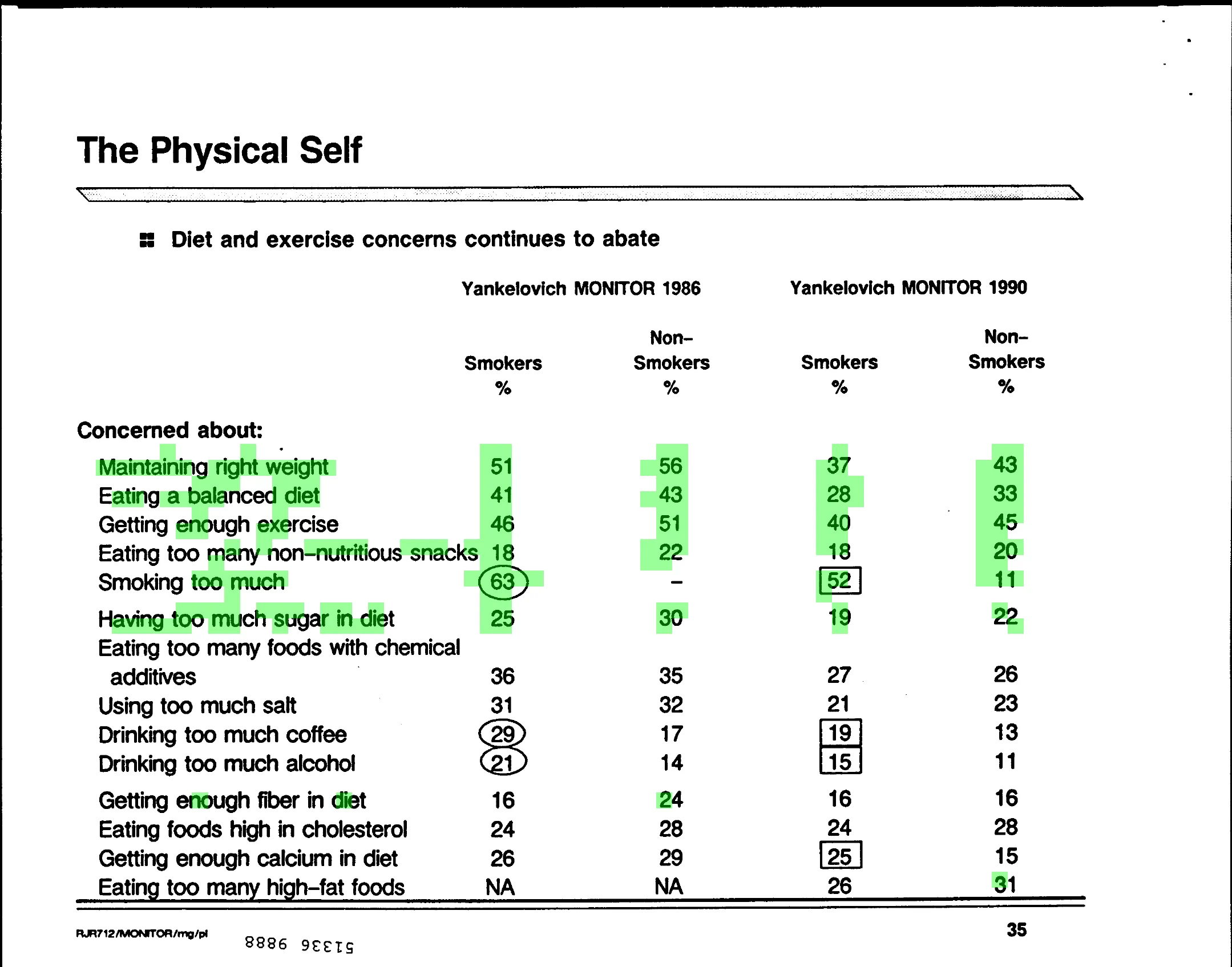}
        \tcblower
        \textbf{Retention ratio:} 3.7\%
        \par\medskip
        \textbf{\Ours{}:} The greatest concern for non-smokers in 1990, according to the chart, is ``\redhighlight{Maintaining right weight},'' with 43\% of non-smokers indicating this as their primary concern.
        \par\medskip
        \textbf{Score:} 0.0
    \end{fullwidthsidebysidebox}
    \caption{
      Failure case 1: The retained visual tokens is insufficient to answer the question correctly.
    }
    \label{fig:failure_case_1}
\end{figure*}

\begin{figure*}[ht]
    \begin{fullwidthsidebysidebox}[righthand width=0.7\textwidth]{myredbg}
        \includegraphics[width=\linewidth]{figures/cases/failure/info_201}
        \tcblower
        \textbf{Question:} How many categories of income mentioned in this infographic?\\
        \textbf{Standard answer:} 6
        \par\medskip
        \textbf{Qwen2.5-VL-7B:} The infographic mentions \redhighlight{five} categories of income:\\

                                1. Under \$25k\\
                                2. \$25-\$35k\\
                                3. \$35-\$50k\\
                                4. \$50-\$75k\\
                                5. Over \$75-\$100k\\
                                \yellowhighlight{6}. Over \$100k\\
        \par\medskip
        \textbf{Score:} 1.0
    \end{fullwidthsidebysidebox}

    \begin{fullwidthsidebysidebox}[righthand width=0.7\textwidth]{mypurplebg}
        \includegraphics[width=\linewidth]{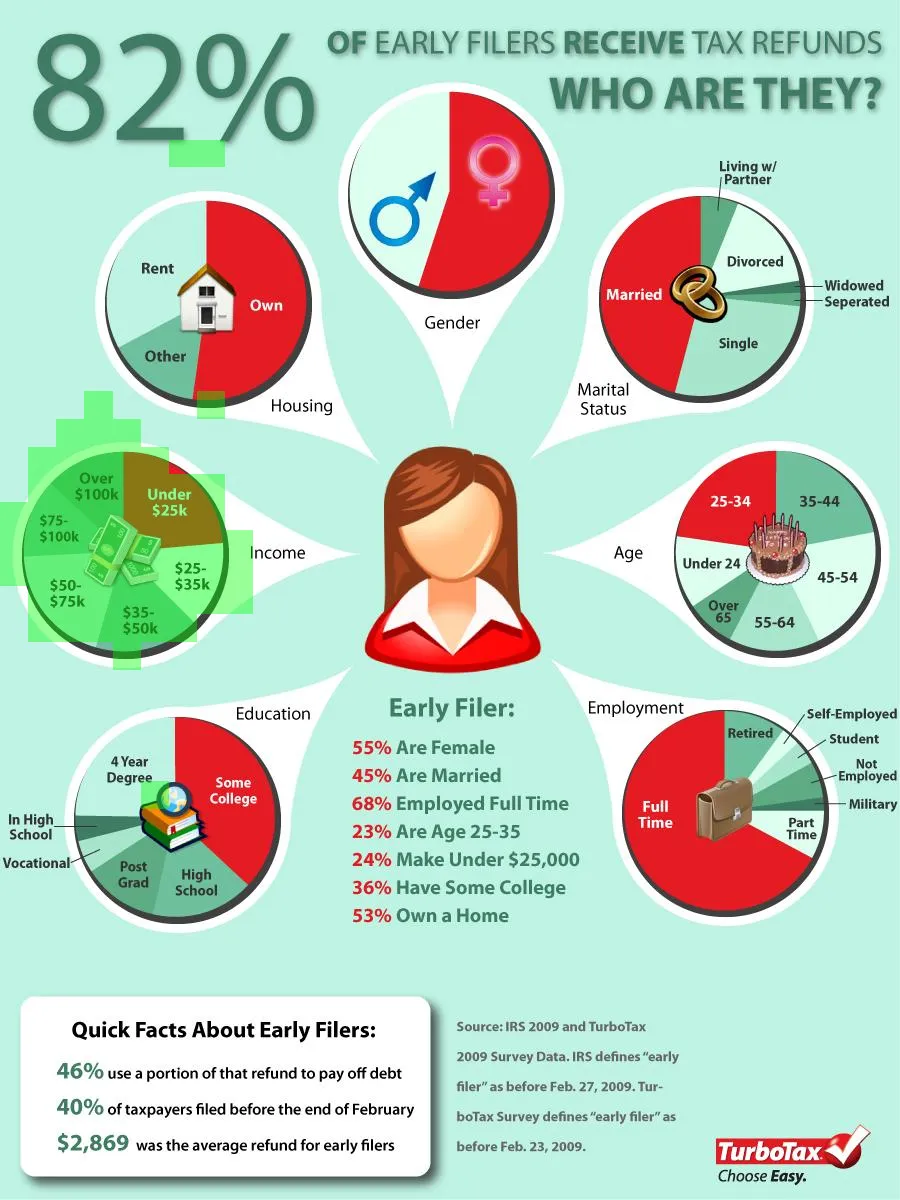}
        \tcblower
        \textbf{Retention ratio:} 4.4\%
        \par\medskip
        \textbf{\Ours{}:} The infographic mentions \redhighlight{five} categories of income:\\

                          1. Under \$25k\\
                          2. \$25-\$35k\\
                          3. \$35-\$50k\\
                          4. \$50-\$75k\\
                          \redhighlight{5}. Over \$100k
        \par\medskip
        \textbf{Score:} 0.8
    \end{fullwidthsidebysidebox}
    \caption{
      Failure case 2: The retained visual tokens is sufficient but the model fails to answer the question correctly.
    }
    \label{fig:failure_case_2}
\end{figure*}

Although \tabref{tab:free_form_vqa} shows that \Ours{} (based on Qwen2.5-VL-7B) matches the baseline's average score with only a 7.4\% retention rate, 
its performance varies across different cases. 
Here, we present both successful and unsuccessful examples to facilitate future research.
\figref{fig:case_1}, \figref{fig:case_2} and \figref{fig:case_3} display successful cases where \Ours{} answers more accurately 
because it prunes question-irrelevant visual information. 
In contrast, \figref{fig:failure_case_1} and \figref{fig:failure_case_2} show two types of failures. 
In the first type (\figref{fig:failure_case_1}), the retained visual information is insufficient. 
In the second type (\figref{fig:failure_case_2}), the information is sufficient, but the model still answers incorrectly. 
We hope these examples provide valuable insights for future studies.

\end{document}